%% file: egpaper_for_review.tex
\documentclass[10pt,twocolumn,letterpaper]{article}

\usepackage{cvpr}
\usepackage{times}
\usepackage{epsfig}
\usepackage{graphicx}
\usepackage{amsmath}
\usepackage{amssymb}
\usepackage{booktabs}
\usepackage{bbm}

\usepackage[pagebackref=true,breaklinks=true,letterpaper=true,colorlinks,bookmarks=false]{hyperref}

\cvprfinalcopy 

\ifcvprfinal\fi
\begin{document}

\title{Image captioning with weakly-supervised attention penalty}

\author{Jiayun Li\\
UCLA\\
{\tt\small jiayunli@g.ucla.edu}
\and
Mohammad K. Ebrahimpour\\
UC Merced\\
{\tt\small mebrahimpour@ucmerced.edu}
\and
Azadeh Moghtaderi\\
Ancestry.com\\
{\tt\small amoghtaderi@ancestry.com}
\and
Yen-Yun Yu\\
Ancestry.com\\
{\tt\small yyu@ancestry.edu}
}

\maketitle

\begin{abstract}
\input{./tex/abstract}
\end{abstract}

\input{./tex/introduction}
\input{./tex/literaturereview}
\input{./tex/model}
\input{./tex/experiment}
\input{./tex/summary}

{\small
\bibliographystyle{ieee}
\bibliography{egbib}
}

\end{document}

%% file: tex/abstract.tex
Stories are essential for genealogy research since they can help build emotional connections with people.
A lot of family stories are reserved in historical photos and albums. Recent development on image captioning models makes it feasible to ``tell stories" for photos automatically. The attention mechanism has been widely adopted in many state-of-the-art encoder-decoder based image captioning models, since it can bridge the gap between the visual part and the language part. Most existing captioning models implicitly trained attention modules with word-likelihood loss. Meanwhile, lots of studies have investigated intrinsic attentions for visual models using gradient-based approaches. Ideally, attention maps predicted by captioning models should be consistent with intrinsic attentions from visual models for any given visual concept. However, no work has been done to align implicitly learned attention maps with intrinsic visual attentions. In this paper, we proposed a novel model that measured consistency between captioning predicted attentions and intrinsic visual attentions. This alignment loss allows explicit attention correction without using any expensive bounding box annotations. We developed and evaluated our model on COCO dataset as well as a genealogical dataset from Ancestry.com Operations Inc., which contains billions of historical photos. The proposed model achieved better performances on all commonly used language evaluation metrics for both datasets. 

%% file: tex/introduction.tex
\section{Introduction}
\label{sec:introduction}
Family historical photo understanding is an open research topic in the genealogical industry and academic community.
In order to discover family history, genealogists start by viewing historical photos to obtain hints and information about family members for constructing pedigrees.
After building the family tree, eventually, genealogists will be able to write down narratives about ancestors.
The entire process is complex and requires lots of human efforts.
In this paper, we proposed a novel model, which can not only generate labels but also descriptions (\eg, captions) for family historical photos.
Image captions can help genealogists efficiently obtain relevant information from photos.
Moreover, image labels obtained from captions can potentially be incorporated into search engines or recommender systems to facilitate capability of genealogy services.
The motivation of this paper is to bridge the gap between genealogy industry and cutting-edge machine learning algorithms, and bring the power of image captioning to the genealogy.

\begin{figure}[t]
\begin{center}
\includegraphics[width=1.1\linewidth]{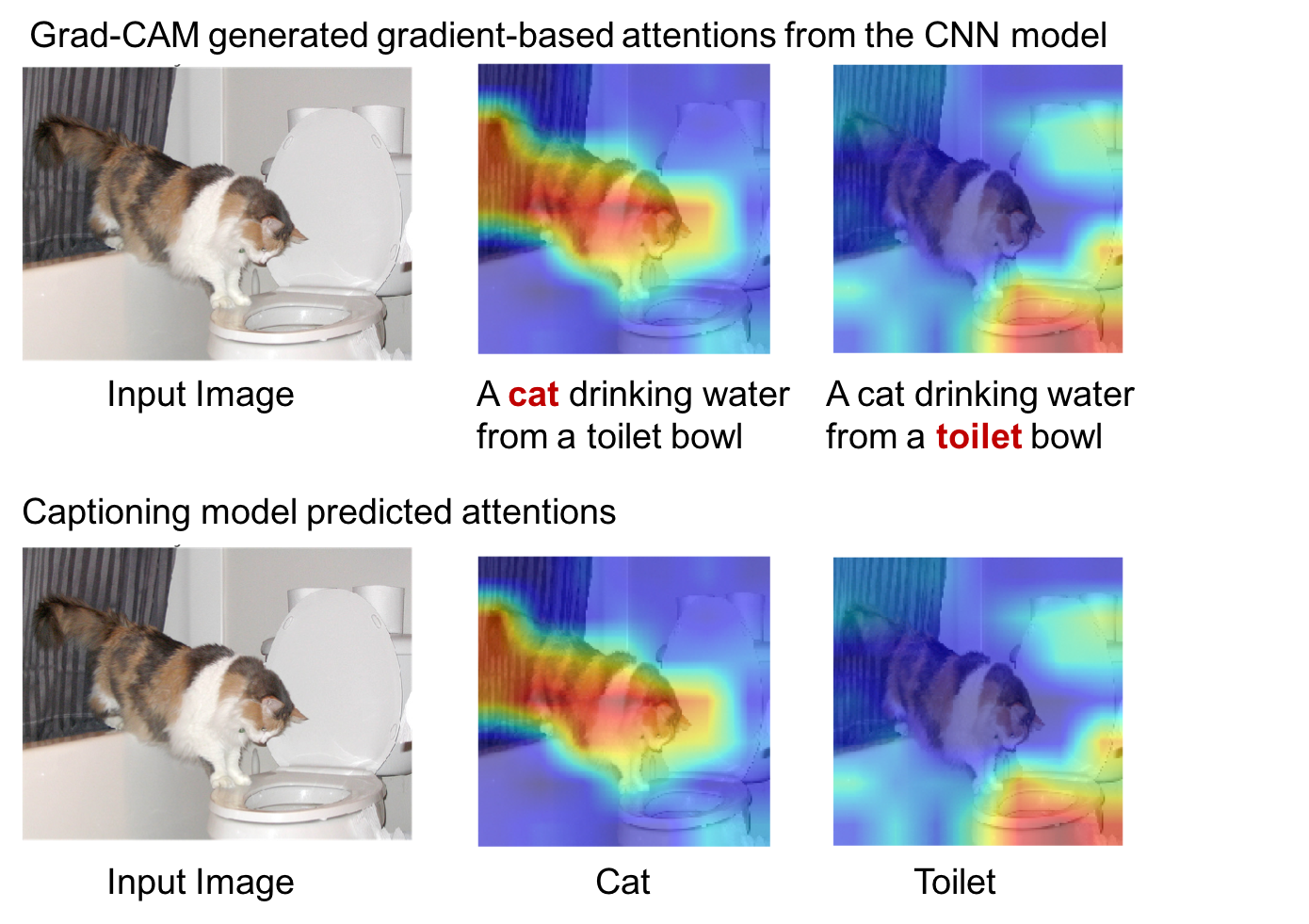}
\end{center}
   \caption{Two types of attention maps: gradient-based intrinsic attentions from the CNN model and attention maps generated by attention modules in the captioning model. Ideally, for every visual word (\eg cat and toilet), these two attention maps should be aligned.}
\label{fig:motivation}
\end{figure}

Automatic image caption generation requires both fully understanding of image contents and the sophisticated generation of natural sentences.
One of the most common approaches is to employ convolutional neural networks (CNN) to encode and represent images into feature vectors followed by long-short term memory (LSTM) models to decode visual information into textual representations \cite{anderson2018bottom, karpathy2015deep, lu2017knowing, lu2018neural, liu2017attention, rennie2017self, wu2016encode, xu2015show}. The attention module in between the CNN and the LSTM helps more relevant visual representations flow into the language model.
Many previous works have explored the attention mechanism,
which not only allows the model to pay attention to important image regions, but also provides visualization about \textit{what} or \textit{where} the model focuses on when predicting the next word \cite{anderson2018bottom, lu2017knowing, lu2018neural, xu2015show}.
An alternative thought is that attention modules provide a sophisticated way of aligning feature representations between the image space and the language space.

Though the attention module is known to be crucial for achieving state-of-the-art performances for captioning models, it's only implicitly trained together with other parts using negative log-likehood loss in most previous works \cite{anderson2018bottom, lu2017knowing, lu2018neural, vinyals2015show, xu2015show}. The lack of explicit attention guidance makes it difficult for captioning models to learn meaningful attention maps, thus leads to inferior performances \cite{liu2017attention}. 
Recently, some state-of-the-art models proposed to use human annotations for attention correction \cite{liu2017attention}.
However, this type of approach requires fine-grained annotations for each visual concept in captions \cite{lu2018neural,liu2017attention}. Fine-grained labels such as bounding-box are available in most open-source captioning datasets, but it can be very expensive to collect for a new dataset.  

In this paper, we proposed a novel image captioning model,
which leveraged attention maps from:
i)  gradients-based weakly-supervised models, such as gradient-weighted class activation mapping (Grad-CAM)
ii) the state-of-the-art bottom-up and top-down attention mechanism in captioning models.
Main contributions of this paper can be summarized as follows: i) We investigated the alignment between attention maps produced by two different mechanisms: captioning attentions and gradient-based CNN attentions. ii) We incorporated Grad-CAM, which has been mainly used for CNN model visualization, to generate saliency maps for each visual concept in the image caption. Specifically, as shown in Figure \ref{fig:motivation} the alignment between intrinsic CNN saliency maps and predicted attentions enables the captioning model to better attend to relevant image regions and ``tell" the sentence according to what has been ``seen". iii) We evaluated and applied the proposed image captioning model on both COCO dataset and Ancestry.com Operations Inc. dataset with historical photos.

%% file: tex/literaturereview.tex
\section{Related Work}
\label{sec:realtedwork}
Image captioning models require joint reasoning in visual and textual space. Prior to encoder-decoder based models, most earlier works in image captioning roughly fell into two main categories: the template-based caption generation and the retrieval-based generation. The first one relied on pre-defined language templates, and then filled in appropriate ``slots" with detected objects class and attributes \cite{elliott2013image, farhadi2010every, kulkarni2013babytalk, li2011composing, mitchell2012midge, yang2011corpus}. The retrieval-based method extracted representations for images and sentences, and retrieved most relevant captions in either the visual space or the multimodal embedding space for visual and textual features \cite{gupta2012choosing, ordonez2011im2text, hodosh2013framing, gong2014improving, karpathy2014deep, bernardi2016automatic}. For example, Karpathy \etal \cite{karpathy2014deep} proposed a model that embedded detected objects and sentence segments into the multimodal feature space, and was trained with multiple instance learning objectives \cite{andrews2002multiple}.

{\noindent \bf Encoder-Decoder based Models.} Different from previous alignment methods, Kiros \etal \cite{kiros2014multimodal, kiros2014unifying} proposed a general encoder-decoder framework for image captioning. The model, which was inspired by the recent development in sequence-to-sequence learning for machine translation \cite{cho2014learning, bahdanau2014neural, sutskever2014sequence, kalchbrenner2013recurrent}, encoded image-text pairs into the common embedding space via a CNN and a LSTM, known as the encoder. The decoder, a neural language model (LSTM), then utilized this joint space to generate words in sequence \cite{bernardi2016automatic, kiros2014unifying}. Visual representations were projected into the LSTM at each time step \cite{bernardi2016automatic, kiros2014unifying}. In contrast, Vinyals \etal \cite{vinyals2015show, vinyals2017show} proposed a model that only fed image features into the LSTM once at the initial step.

{\noindent \bf Attention Matters.} These encoder-decoder based models achieved promising performances and gained a lot of popularity \cite{donahue2015long, mao2014deep, vinyals2015show, vinyals2017show}. However, they didn't take into account of the fact that the network may attend to different parts of images changing when generating different words. Attention plays an important role in human visual system, since it enables brain to focus the most important information when dealing with the complex signals in nature \cite{rasolzadeh2007attentional, desimone1995neural}. Similarly, instead of processing information from the entire input, the attention mechanism allows deep neural networks to pay attention to parts that are most relevant for solving the current problem, and it has been shown to be one of key components for achieving state-of-the-art performances in many machine learning tasks \cite{anderson2018bottom, bahdanau2014neural, chen2016attention, liu2017attention, lu2017knowing, lu2018neural, vaswani2017attention, xu2015show}. Besides improving predictions of neural networks, the attention approach can also provide visual explanations for network decisions by highlighting important regions corresponding to any decision of interest \cite{selvaraju2017grad, zhou2016learning}.

{\noindent \bf Attention Mechanisms in Image Captioning.} Recently, attention modules have been embedded in many encoder-decoder-based captioning models in order to selectively incorporate visual context \cite{anderson2018bottom, karpathy2015deep, lu2017knowing, lu2018neural, liu2017attention, rennie2017self, wu2016encode, xu2015show}. For example, Xu \etal \cite{xu2015show} proposed two different attention approaches: the stochastic ``hard" attention and the determinstic ``soft" attention to encourage the captioning model to pay more attention to image parts, which are relevant to each generated word. In the adaptive attention model \cite{lu2017knowing}, a ``visual sentinel" was utilized to decide whether the decoder should more rely on visual information or textual context when predicting the next word. Attention modules were further improved by using a top-down and bottom-up approach \cite{anderson2018bottom}. Attention mechanisms bridge the gap between the visual world (CNN features) and the language world (LSTM predictions) \cite{liu2017attention}. Despite many previous works have explored how the attention mechanism can help networks make better predictions, as well as how it can visualize the ``thinking process" of neural networks, few work has been done to quantitatively evaluate attention maps predicted by models.
Liu \etal \cite{liu2017attention} investigated correctness of attention maps predicted by LSTM-CNN models and incorporated additional supervision by aligning predicted attentions with human annotations. However, this approach requires bounding boxes annotations. To the best of our knowledge there is no work on utilizing gradients-based attention maps as weak supervision and measuring the consistency between intrinsic CNN attentions and LSTM predicted attentions. In this paper, we proposed a weakly-supervised attention alignment model that penalized discrepancies between the attention map predicted by the captioning model and the one generated by gradients from the CNN model. This is also the first captioning model that has been developed and applied on the historical dataset. The proposed attention alignment model provided additional supervision for attention modules in LSTM and showed improvements in commonly used language evaluation metrics.

%% file: tex/model.tex
\section{Method}
\label{sec:model}
Most state-of-the-art image captioning models contain both a language model (\eg , LSTM) and a visual model (\eg , CNN), and follow the ``encoder-decoder" architecture. Instead of encoding the entire image into one static representation, the attention mechanism, which is shown to be very effective in captioning models, allows the network to ``look at" different parts of the image when generating different words. Most previous works utilized image features from a pre-trained CNN (\eg Resnet \cite{he2016deep}, VGG \cite{simonyan2014very}) and hidden state of the LSTM $\boldsymbol{h}_t$ to predict which region to be focused on at each step $t$. If we denote parameters in the captioning model as $\boldsymbol{\theta}$ and words generated in the previous $t-1$ steps as $\boldsymbol{y}_{1:t-1}$, the overall objective of most captioning models can be defined in Eq (\ref{eq:mle}). However, attention maps of these methods were only implicitly trained with negative log likelihood loss from words as defined in Eq (\ref{eq:mle}). More importantly, they didn't take into account of intrinsic saliency maps from the visual model.
\begin{align}\label{eq:mle}
\begin{split}
\boldsymbol{\theta^{\star}} &=-\mathrm{argmin}_{\boldsymbol{\theta}}\sum_{t}{\mathrm{log}(p(\boldsymbol{y}_t|\boldsymbol{y}_{1:t-1}; \boldsymbol{\theta}))}
\end{split}
\end{align}
Different from existing works, our proposed model incorporates gradient-based attention information from the visual model by aligning the intrinsic attention with the predicted attention. In section \ref{subsec:model:general_att}, we will describe general attention approaches in most image captioning models, then we will discuss how to utilize state-of-the-art visualization techniques of CNN models to produce weakly-supervised attention maps in section \ref{subsec:model:grad_att}. In section \ref{subsec:model:dual_att}, we will demonstrate how our weakly-supervised attention correction model utilizes both types of attention maps.
\subsection{Attention Mechanism in Captioning Models}
\label{subsec:model:general_att}
State-of-the-art CNN models (\eg , ResNet 101 \cite{he2016deep}) are embedded in most image captioning models to extract visual representations for a given image $\boldsymbol{I}$. Visual features of $k$ image regions from the last convolutional layer before the fully connected layer can be flattened and represented as a set of feature vectors $\boldsymbol{V} \in \mathbb{R}^{k \times d}$ as defined in Eq (\ref{eq:feature_vector}), each of which is a $\mathrm{d}$ dimensional feature vector $\boldsymbol{v}_i \in \mathbb{R}^{d}$ corresponding to one region of the image. For example, we assume the size of feature maps from the last convolutional layer is $512 \times 7 \times 7$. $512$ is the number of filters. Then these feature maps can be flattened into $512 \times 49$ features, which corresponds to $\mathrm{d} = 512$ dimensional feature vectors from $k = 49$ image regions. 
\begin{align}\label{eq:feature_vector}
\boldsymbol{V} := \{\mathbf{v}_i\}_{i=1}^{k}, \textrm{where}\ \boldsymbol{v}_i \in \mathbb{R}^{d}
\end{align}

\begin{figure}[t]
\begin{center}
\includegraphics[width=1.0\linewidth]{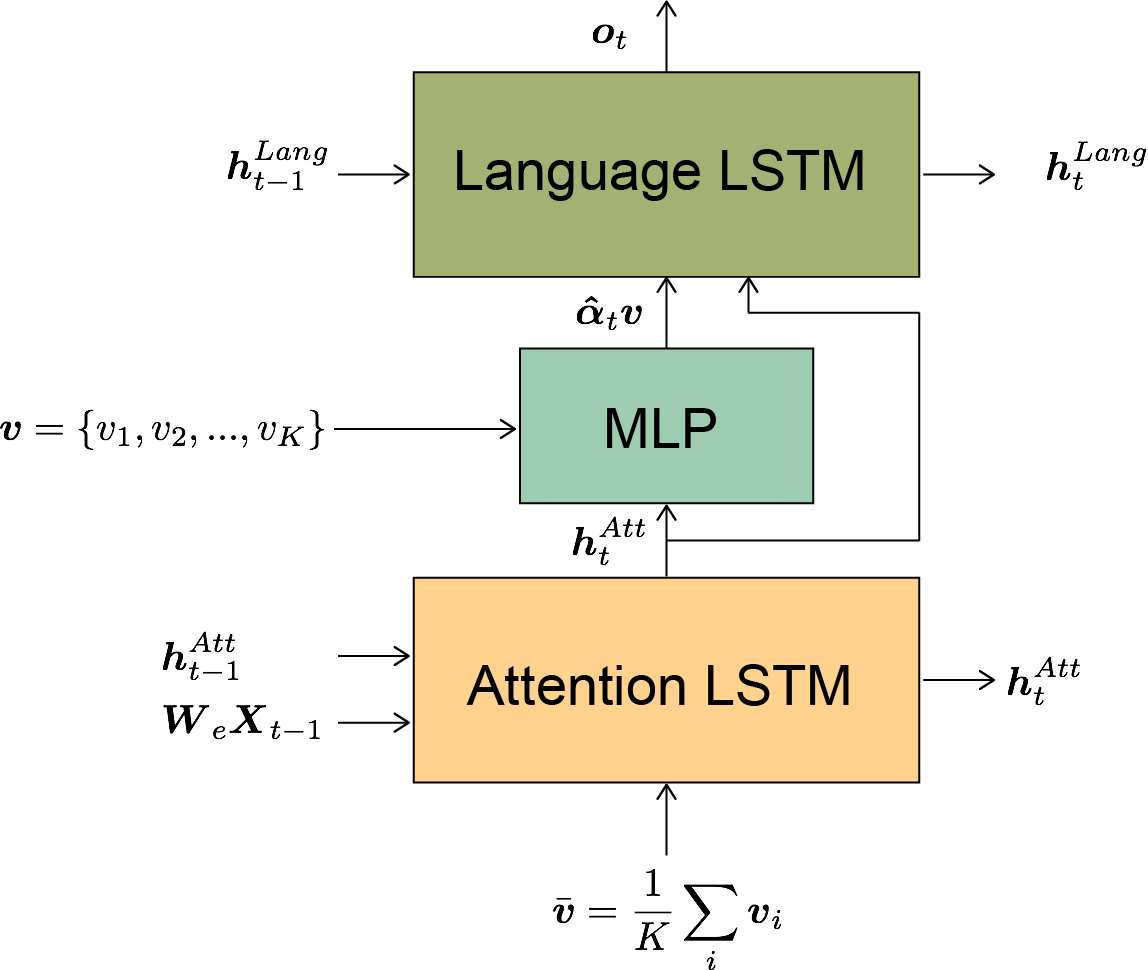}
\end{center}
   \caption{The attention mechanism in the caption model, which utilized both top-down attention from average pooling features $\bar{\boldsymbol{v}}$ and bottom-up attention from visual representations of different regions $\boldsymbol{V}$.}
\label{figure:att_cap_fig}
\end{figure}

If we denote hidden states of LSTM as $\boldsymbol{h} \in \mathbb{R}^{n}$, the attention map $\boldsymbol{\hat{\alpha}}_t$ at each time step $t$ can be determined by different types of attention models $f_{att}(\boldsymbol{v}, \boldsymbol{h})$. $f(\cdot)$ can be multi-layer perceptron (MLP) \cite{liu2017attention, lu2017knowing, xu2015show} or LSTM \cite{anderson2018bottom, lu2018neural}. Concretely, in MLP-based attention models, spatial image features combined with hidden states of LSTM at previous time step $\boldsymbol{h}_{t-1}$ \cite{xu2015show} or the current step $\boldsymbol{h}_t$ \cite{lu2017knowing} are fed into a fully connected layer followed by a softmax layer to generate attention distributions over $k$ regions. For example, the attention module conditioned on the previous time step $\boldsymbol{h}_{t-1}$ can be defined in Eq (\ref{eq:cap_att}).
\begin{align}\label{eq:cap_att}
\hat{\boldsymbol{\alpha}}_t &= \textrm{softmax}[\boldsymbol{\omega}_c^{T}\mathrm{tanh}(\mathbf{W}_v\boldsymbol{V} + \mathbf{W}_h\boldsymbol{h}_{t-1}\mathbbm{1}^{T})]
\end{align}
where $\mathbf{W}_v \in \mathbb{R}^{k \times d}, \mathbf{W}_h \in \mathbb{R}^{k \times n}$ and $\boldsymbol{\omega}_c \in \mathbb{R}^k$ are trainable parameters in MLP. $\mathbbm{1} \in \mathbb{R}^{k}$ is a vector with all elements that equal to 1. Attention modules can also be modeled with a LSTM \cite{anderson2018bottom, lu2018neural}, which takes previous hidden states $\boldsymbol{h}_{t-1}$, the embedding of previous generated word $\mathbf{W}_e\mathbf{X}$, and the average pooled visual features $\bar{\boldsymbol{v}} = \frac{1}{K}\sum_i{\boldsymbol{v}_i}$ as inputs. Outputs from the attention LSTM, $\boldsymbol{h}_{att}$ are fed into a fully connected layer and a softmax layer to predict attention weights, as defined in Eq (\ref{eq:cap_att_lstm}). 
\begin{align}\label{eq:cap_att_lstm}
  \begin{split}
    \boldsymbol{h}_{att} &= \mathrm{LSTM}(\boldsymbol{h}_{t-1}, \bar{\boldsymbol{v}}, \mathbf{W}_e^{T}\mathbf{X}_{t-1}) \\
    \hat{\boldsymbol{\alpha}}_t &= \textrm{softmax}[\mathbf{\omega}_c^{T}\mathrm{tanh}(\mathbf{W}_v\boldsymbol{V} + \mathbf{W}_h\boldsymbol{h}_{att}\mathbbm{1}^{T})]
  \end{split}
\end{align}
where $\mathbf{X} \in \mathbb{R}^{m \times m}$ is the one-hot representation for the vocabulary of size $m$. $\mathbf{W}_e \in \mathbb{R} ^{m \times e}$ is parameters for $e$ dimensional word embedding. This approach combines both bottom-up and top-down features and has been shown to be very effective \cite{anderson2018bottom}.
In this paper, we adopt the similar attention architecture as \cite{anderson2018bottom, lu2018neural} shown in Figure \ref{figure:att_cap_fig}. Since our approach doesn't require fine-grained labels, bottom-up features are obtained from grid-based image regions instead of object regions defined by bounding-box annotations. 

\subsection{Gradient-based Attention Method}
\label{subsec:model:grad_att}
In section \ref{subsec:model:general_att}, we briefly describe most commonly used attention methods in image captioning models. In this section \ref{subsec:model:grad_att}, we will present how to generate ``visual attentions" from the CNN, which will be incorporated in the attention penalty. The CNN model, known as the encoder in image captioning models, can not only provide rich visual representations, but also generate weakly-supervised localization of discriminative image regions for a given target concept. Grad-CAM \cite{selvaraju2017grad}, for instance, can be utilized to produce ``visual explanation maps", which may help understand the decision making process of deep neural networks. Given a trained CNN model and a target class $c$, in order to visualize discriminative regions, we can first compute network outputs $\boldsymbol{o}_c$ before the softmax layer for class $c$. Then gradients of $\boldsymbol{o}_c$ \wrt activations $\mathbf{A}^{l}$ of $l$-th feature map in the convolutional layer will be obtained through backpropagation. Global average pooling over $k$ regions can be used to generate weights that represent the importance of $w \times h$ feature maps. Weighted combination of $d$ dimensional feature maps will then determine the attention distribution of $k$ regions\cite{selvaraju2017grad} for predicting the target class $c$ as defined in Eq (\ref{eq:grad_cam}).

\begin{align} \label{eq:grad_cam}
  \begin{split}
    \boldsymbol{\theta}_l^c &= \frac{1}{Z}\sum_{i \in w}\sum_{j \in h}{\frac{\partial{\boldsymbol{o}_c}}{\partial{\mathbf{A}^l_{i,j}}}} \\
    \boldsymbol{\alpha}^c &= \mathrm{ReLU}(\sum_{l=1}^d{\boldsymbol{\theta}_l^c\mathbf{A}^l})
  \end{split}
\end{align}
where $Z = w \times h$ is the normalization constant.
The Grad-CAM produces a coarse heatmap of $c$-th class, $\boldsymbol{\alpha}^c$, which has the same spatial size as feature maps of the convolutional layer $i$ (\eg, if $i$ is the last convolutional layer of the ResNet\cite{he2016deep}, the size of $\boldsymbol{\alpha}^c$ will be $7\times7$). The ReLU function removes the effect of pixels with negative weights, since they don't have positive influence in predicting current class.   

\begin{figure}[h]
\begin{center}
\includegraphics[width=1.0\linewidth]{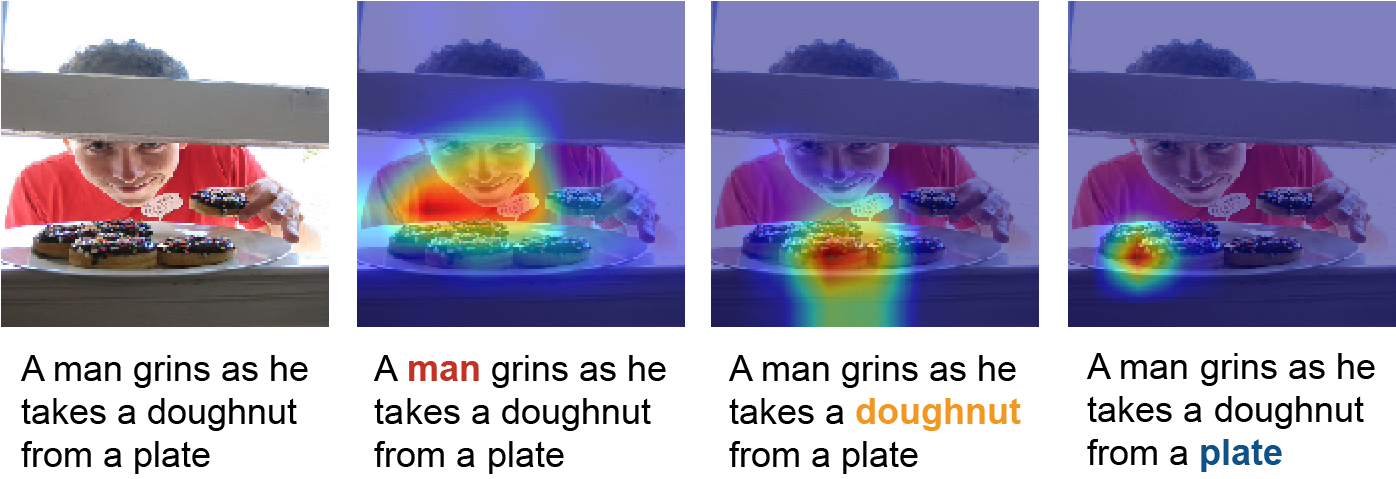}
\end{center}
   \caption{Attention maps generated for each word (bold and annotated with colors) in the caption. The pre-trained CNN is fine-tuned with labels generated by parsing captions and mapping visual concepts to object categories. For each visual word, the attention map is produced by computing gradients of CNN outputs \wrt the last convolutional layer.}
\label{figure:att_grad_cam}
\end{figure}

Grad-CAM is able to highlight saliency image parts for various types of neural networks, such as image captioning models. Concretely, attention maps for a given caption were generated by computing gradients of its' log likelihood \wrt the last convolutional layer of the encoder \cite{selvaraju2017grad}. However, these attention maps are static at different time steps. If the caption contains multiple visual concepts, the model will identify discriminative regions for all targets at once. For example, given the caption that ``A man grins as he takes a doughnut from a plate" as shown in Figure \ref{figure:att_grad_cam}, backpropogating its' log likelihood will produce an attention map highlighting ``man" ``doughnut" and ``plate".

In order to generate attention maps dynamically with Grad-CAM, we first parse the caption using Stanford CoreNLP toolkit \cite{manning2014stanford} to get words that may be related to visual concepts (\eg , Nouns). A word to object categories mapping dictionary, which maps specific visual words to general object concepts (\eg , map word ``kitten" to ``cat" category), can either be manually built \cite{lu2018neural} or automatically constructed by measuring cosine similarity between the word and the object category in the word embedding space \cite{liu2017attention}. Then the image will have one or multiple labels according to object categories mentioned in captions. After that, the CNN model pre-trained on ImageNet will be fine-tuned using the multi-label multi-class classification objective. Instead of computing gradients of negative log probability of a caption, now we can calculate gradients of class scores for a given visual word. Hence, different attention maps can be generated for different visual concepts in the caption as shown in Figure \ref{figure:att_grad_cam}. Note that ``man" ``doughnut" and ``plate" are highlighted individually in different attention maps. 

\subsection{Weakly-supervised Attention Alignment}
\label{subsec:model:dual_att}
The captioning-based attention maps as discussed in section \ref{subsec:model:general_att} are usually only trained with negative log likelihood loss of generated word sequences, and there is no explicit loss for attention maps. Previous work \cite{liu2017attention} has shown that adding supervised additional penalty in the loss function can improve quality of both generated attention maps and captions. The supervised attention penalty provides a way to quantize the consistency between the predicted attention and bounding boxes annotations of Noun entities. 

However, it can be very expensive and time-consuming to get a large amount of bounding box annotations for a new dataset. The proposed weakly-supervised attention alignment model computes the differences between predicted and gradient-based saliency maps without requiring for any manually fine-grained labels. The attention module uses visual features from the CNN model and outputs from the LSTM to predict the captioning model should ``pay attention to" which part of the image to when generating the current word $X_t$. Ideally, for any given visual concept, the predicted attention distribution should align with the gradient-based CNN attention map.

\begin{figure*}[h]
\begin{center}
\includegraphics[width=1.0\linewidth]{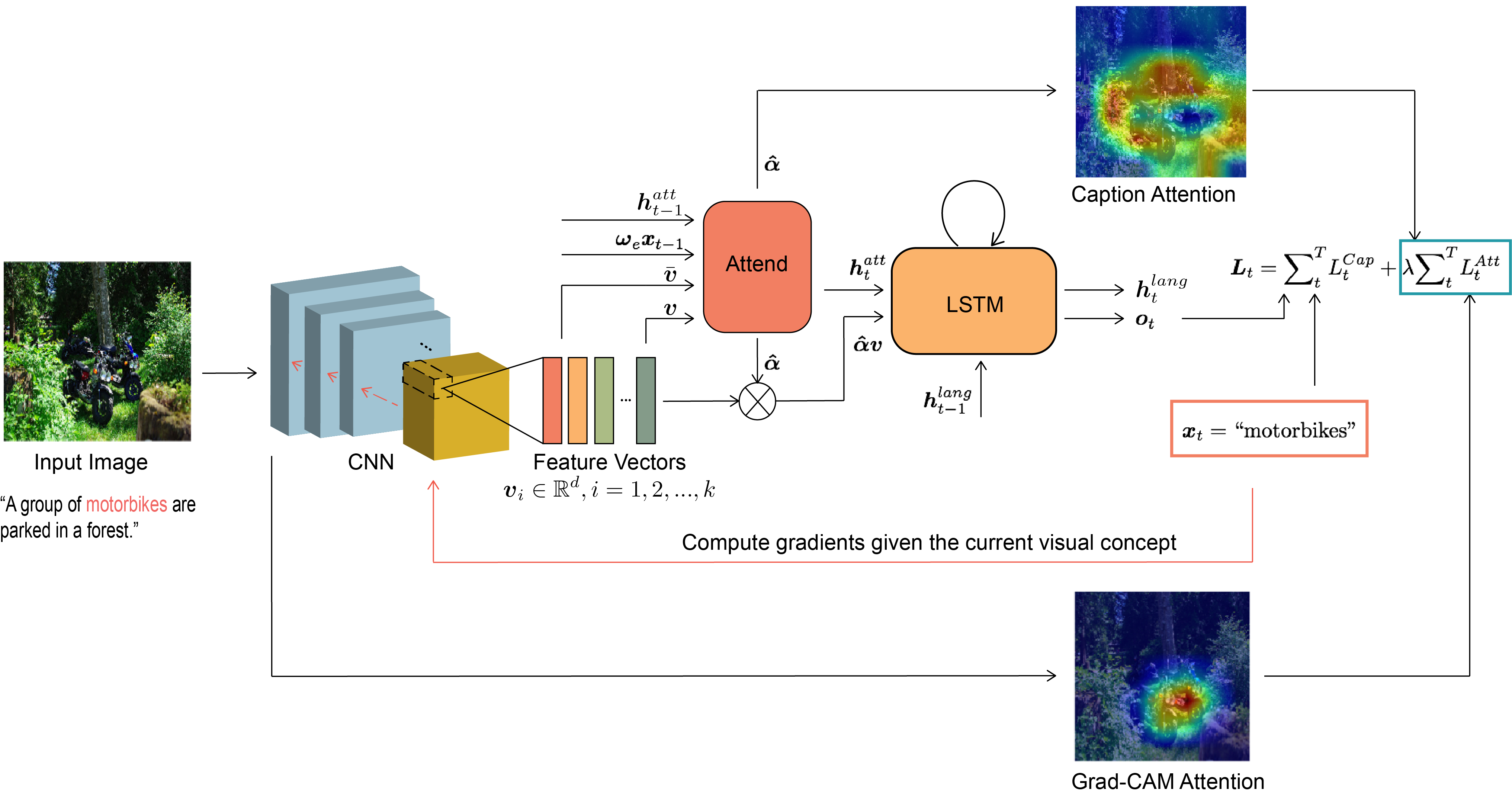}
\end{center}
   \caption{Overview of the proposed attention alignment model. The model can be roughly divided into three parts: a visual part (CNN), an attention module (Attend) and a language part (LSTM). The CNN model will encode the image into $d \times w \times h$ feature maps, which can be flatten into $k$ feature vectors corresponding to $k$ regions. The attention module uses bottom-up top-down features and embeddings of previously generated word to predict relevant areas for generating current word. The LSTM takes weighted feature maps, previous hidden states as inputs and predict current word. In additional to the loss from captioning model, we compute the cross entropy loss between predicted and intrinsic gradient-based attention maps. }
\label{fig:dual_att_overview}
\end{figure*}

During training, in order to predict the current word the captioning model uses the textual context from the word embedding of the previous word and previous LSTM hidden states, as well as visual context from weighted image features. These weights $\boldsymbol{\hat{\alpha}}$ that obtained from the attention module $f_{att}$ essentially approximate the importance of image parts when generating the current word. Meanwhile, the gradient-based attention distribution $\boldsymbol{\alpha}$ from Grad-CAM highlights the relevant regions given the ground truth of the current word $X_t$. Following \cite{liu2017attention}, We can consider $\boldsymbol{\hat{\alpha}} \in \mathbb{R}^{k}$ and $\boldsymbol{\alpha} \in \mathbb{R}^{k}$ as two probability distributions, and use cross entropy loss to measure discrepancies between two distributions. If the current word is not Noun or not included in the object category dictionary, the attention penalty is set to be 0, since the loss can be ambiguous to define. The training objective of our model is summarized in Eq (\ref{eq:loss_att}) and Eq (\ref{eq:loss_tot}).  Figure \ref{fig:dual_att_overview} shows an overview of our model architecture.

\begin{align}\label{eq:loss_att}
    \boldsymbol{L}^{Att}_t &=
    \begin{cases}
      -\sum_{i=1}^{K}{\boldsymbol{\alpha}_{ti}\mathrm{log}\boldsymbol{\hat{\alpha}}_{ti}},& \text{if } \boldsymbol{\alpha}_{t} \text{ exists for } x_t\\
      0,              & \text{otherwise}
    \end{cases}
\end{align}

\begin{align} \label{eq:loss_tot}
  \begin{split}
  \boldsymbol{L}_{t} &= \boldsymbol\sum^{T}_t{{L}^{Cap}_t} + \lambda\boldsymbol\sum^{T}_t{{L}^{Att}_t} \\
  \boldsymbol{L}^{Cap}_t &= -\mathrm{log}[p(\boldsymbol{y_t}|\boldsymbol{y}_{1:t-1}; \boldsymbol{\theta})]
  \end{split}
\end{align}

%% file: tex/experiment.tex
\section{Experiments}
\label{sec:experiments}
In section \ref{subsec:experiments:settings}, we will introduce evaluation metrics and two datasets that have been used for our model development and evaluation. We provide more details about model implementation in section \ref{subsec:experiments:implementation}. Results on COCO dataset and Ancestry.com Operations Inc. dataset are presented in section \ref{subsec:experiments:coco} and section \ref{subsec:experiments:ugc}, respectively. 
\subsection{Experiment Settings}
\label{subsec:experiments:settings}

We experimented with two different datasets: COCO dataset and Ancestry.com Operations Inc. dataset containing historical photos.

{\noindent \bf COCO} is a large-scale dataset for image captioning, segmentation, classification and object detection, containing 82,783 images for training, 40,504 for validation and 40,775 for testing \cite{lin2014microsoft}. Each image is annotated by 5 crowdsourced captions. We used the same train/val splits as in \cite{karpathy2015deep, lu2017knowing, lu2018neural, liu2017attention} containing 5000 images for validation and 5000 for testing. We utilized the Stanford CoreNLP \cite{manning2014stanford} to parse captions into tokens, and built a vocabulary by using words that appeared more than 5 times in the dataset. The maximum length of captions was set to be 20.

{\noindent \bf Ancestry.com Operations Inc. dataset} is one of the largest genealogy  dataset that contains around billions of digitized historical photos uploaded by users. We collected 2 crowdsourced captions per image for around 10,000 images. Human annotators were trained to follow the similar annotation protocol as in \cite{lin2014microsoft}, and the annotation was performed using the in-house labeling tool. The distinction between this dataset and the open source dataset in terms of image quality, data domain, and \etc, makes it challenging to directly apply off-the-shelf captioning model developed on modern datasets. For instance, historical images may contain objects that are less common in modern datasets (\eg, horse-drawn carriages are quite common in historical photos but not in modern pictures).  We used the same preprocessing pipeline as on COCO dataset to build the vocabulary for Ancestry.com Operations Inc. dataset. Since the size of this dataset is relatively small, we transferred the model trained on COCO dataset and fine-tuned it on Ancestry.com Operations Inc. dataset.

{\noindent \bf Evaluation.} We utilized the publicly available COCO-captioning evaluation toolkit \cite{lin2014microsoft}, which computed widely used automatic language evaluation metrics: BLEU \cite{papineni2002bleu}, METEOR \cite{denkowski2014meteor}, CIDEr \cite{vedantam2015cider}, ROUGE-L \cite{lin2004rouge}, and SPICE \cite{anderson2016spice}. We reported model performances on test splits of COCO dataset and Ancestry.com Operations Inc. dataset.

\subsection{Implementation Details}
\label{subsec:experiments:implementation}
{\noindent \bf Multi-label Classification Model.} We used a manually built word to object mapping as developed by \cite{lu2018neural}, which has mappings between 413 fine-grained classes and 80 object categories. We adopted the ResNet-101 \cite{he2016deep} that has been pre-trained on ImageNet \cite{krizhevsky2012imagenet} as the feature extractor in the captioning model. To fine-tune the model on COCO dataset \cite{lin2014microsoft}, as described in section \ref{subsec:model:grad_att}, instead of relying on bounding boxes, visual concepts generated from image captions were used as labels for classification models. The fully connected layer with 1000 output neurons was replaced by the one with 80 outputs corresponding to 80 object categories in COCO dataset. The fully connected layer was trained with an initial learning rate at 0.0001 for 50 epochs. Since shallow layers in CNN models tend to learn more general representations, we fine-tuned the whole model except the first two convolutional blocks for ResNet-101 model with an initial learning rate at 0.00001. Adam \cite{kingma2014adam} was used for model optimization. 

{\noindent \bf Image caption model with attention penalty.}
The attention penalty was computed by aligning the LSTM predicted attention with CNN-based saliency maps for every visual concept in captions. Specifically, the category of each visual word was backpropagated to the last convolutional layer of the trained feature extractor (ResNet-101 \cite{he2016deep}) to generate class specific attention maps \cite{selvaraju2017grad}. We used the word embedding size of 512. The hidden size of attention LSTM and language LSTM was set at 1024. To incorporate attention alignment penalty into the final loss function, we experimented with both adaptive and constant ${\lambda}$, and the constant $\lambda$ at 100 gave the best performance. We used the Adam optimizer and an initial learning rate starts at 0.0005 for captioning model optimization. The scheduled sampling was used to reduce differences between sentence generation during training and inference \cite{bengio2015scheduled}. Beam size was set at 3 for model evaluation.

\begin{table*}[h]
\begin{center}\caption{Model performances on the test split of COCO dataset}
\label{tab:coco_performance}
\begin{tabular}{@{} l *8c @{}}
\toprule
 \multicolumn{1}{c}{Methods}  & BLEU-1 & BLEU-2 & BLEU-3 & BLEU-4 & METEOR & CIDEr & SPICE & ROUGE-L\\
 \midrule
 Top-down and Bottom-up \cite{anderson2018bottom}  & 77.2 & - & - & 36.2 & 27.0 & 113.5 & 20.3 & 56.4  \\
 Neural Baby Talk \cite{lu2018neural} & 75.5 & - & - & 34.7 & 27.1 & 107.2 & 20.1 & - \\
 \midrule
 Google NIC \cite{vinyals2015show} & 66.6 & 46.1 & 32.9 & 24.6 & - & - & - & - \\
 Soft Attention \cite{xu2015show} & 70.7 & 49.2 & 34.4 & 24.3 & 23.9 & - & - & - \\
 Hard Attention \cite{xu2015show} & 71.8 & 50.4 & 35.7 & 25.0 & 23.0 & - & - & - \\
 \midrule
 Attention Correctness \cite{liu2017attention} & - & - & 37.2 & 27.6 & 24.8 & - & - & - \\
 Ours ablated & 71.8 & 54.9 & 41.4 & 31.3 & 24.6 & 95.1 & 17.6 & 52.8 \\
 Ours w attention & \bf{72.0} & \bf{55.1} & \bf{41.5} & \bf{31.4} & \bf{24.7} & \bf{95.6} & \bf{17.7} & \bf{53.0} \\
 \bottomrule
\end{tabular}
\end{center} \label{tab:coco_data}
\end{table*}  

\begin{figure*}[t]
\begin{center}
\includegraphics[width=0.9\linewidth]{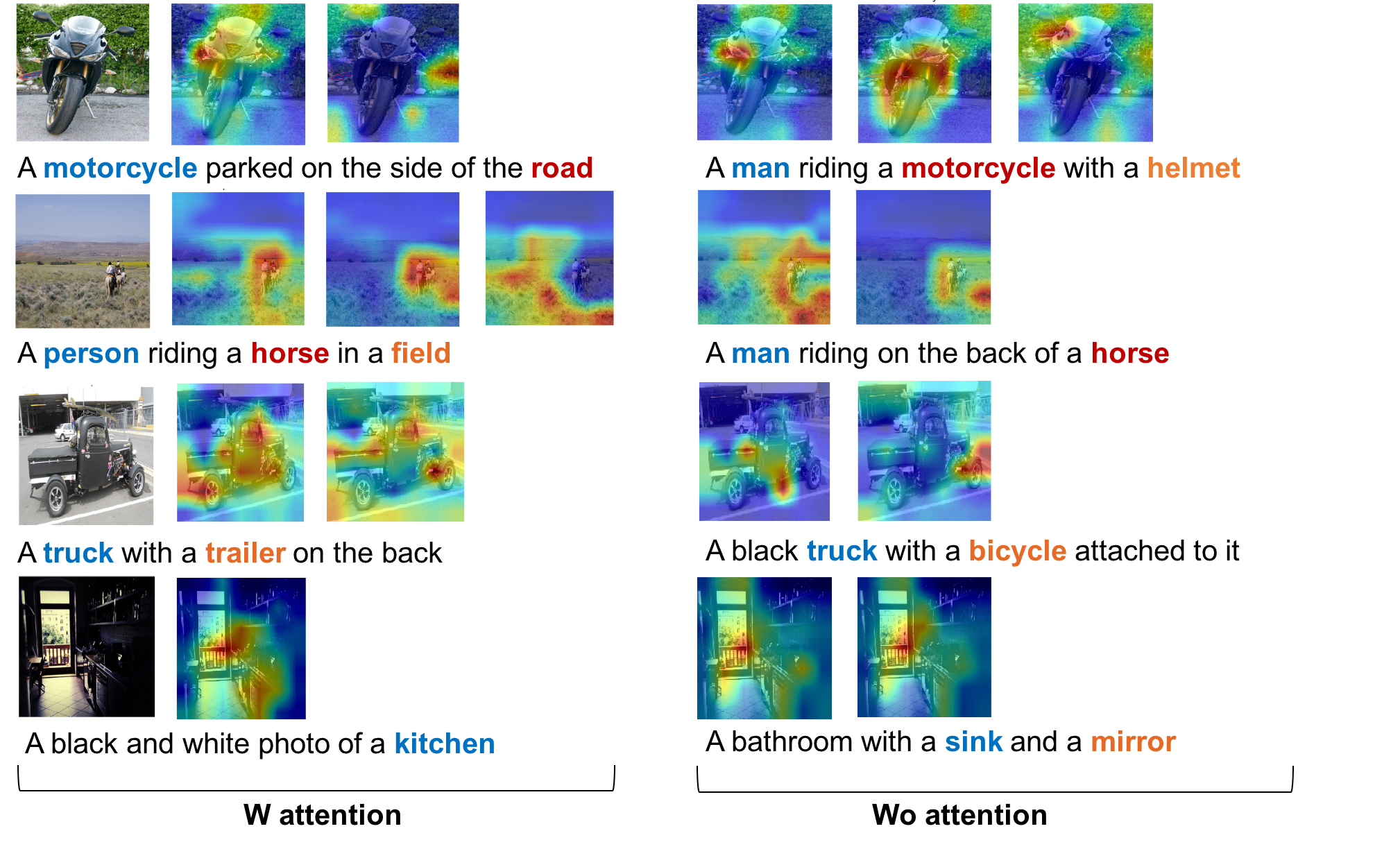}
\end{center}
   \caption{Sample results on COCO dataset for both proposed model (w attention) and the ablated one (wo attention). }
\label{fig:sample_result_coco}
\end{figure*}
\subsection{Results: COCO}
\label{subsec:experiments:coco}
We compared our method with captioning models that don't rely one bounding box annotations: Google NIC \cite{vinyals2015show}, Soft Attention \cite{xu2015show} and Hard Attention \cite{xu2015show}. We also compared with \cite{liu2017attention}, which utilized bounding box annotations as the additional attention supervision. Table \ref{tab:coco_performance} summarized model performances. To evaluate the effect of the attention penalty, we used the same pre-trained CNN model and the LSTM architecture for the ablated model. As shown in Table \ref{tab:coco_performance}, the performance of the proposed with attention model outperformed the ablated model (without attention penalty) on all evaluation metrics. Neural baby talk \cite{lu2018neural} and top-down and bottom-up \cite{anderson2018bottom} achieved better performances, but both models used bounding box annotations that can be difficult to generate for a new dataset (\eg,Ancestry.com Operations Inc. historical dataset). 

Figure \ref{fig:sample_result_coco} shows some sample results on the COCO dataset. We can see that if the predicted attention maps becomes less meaningful, the captioning model may produce sentences that are less grounded in the image. For example, the left side of the second row in Figure \ref{fig:sample_result_coco} shows a motorcycle. However, the attention map generated by the ablated model failed to highlight relevant regions, and generated words such as ``man" and ``helmet" were not grounded in the image. The additional attention alignment loss can help the captioning model to focus on relevant regions during the sentence generation. 

\subsection{Results: Historical Photo}
\label{subsec:experiments:ugc}

\begin{table*}[t]\caption{Model performances on the test split of Ancestry.com Operations Inc. dataset}
\label{tab:ugc_performance}
\begin{center}
\begin{tabular}{@{} l *8c @{}}
\toprule
 \multicolumn{1}{c}{Methods}  & BLEU-1 & BLEU-2 & BLEU-3 & BLEU-4 & METEOR & CIDEr & SPICE & ROUGE-L\\
 \midrule
 Ours ablated & 70.6 & 61.9 & 55.0 & 48.8 & 32.5 & 187.4 & 39.1 & 66.8 \\
 Ours w attention & \bf{72.3} & \bf{64.6} & \bf{58.4} & \bf{52.8} & \bf{35.6} & \bf{233.8} & \bf{44.1} & \bf{68.6} \\
 \bottomrule
\end{tabular}
\end{center}
\end{table*}

Since Ancestry.com Operations Inc. dataset is relatively small compared with COCO dataset, we didn't train the captioning model from the scratch. Instead, layers in the LSTM and attention modules of pre-trained models were fine-tuned on Ancestry.com Operations Inc. dataset. We reported the same evaluation metrics. As shown in Table \ref{tab:ugc_performance}, the proposed attention alignment model achieved much better performances on all evaluation metrics compared with the ablated version. 

%% file: tex/summary.tex
\section{Discussion and Future Work}
\label{sec:summary}
In this paper, we investigated the consistency between gradient-based saliency maps from the CNN model and the predicted attention maps from attention modules in most encoder-decoder based captioning models. We proposed a novel approach that could guide implicitly learned attention maps without relying on fine-grained annotations. The proposed model were evaluated on two distinct datasets and demonstrated better performances across all commonly used evaluation metrics.   

However, the improvement is relatively small on COCO dataset, but large on Ancestry.com Operations Inc. historical dataset. This could be caused by the fact that both attention maps haven't been trained with strong supervision, thus may not be perfect. Most historical photos in Ancestry.com Operations Inc. dataset were taken with the purpose of recording the important events or moments, so relevant contents (\eg people) may usually be large and centered, which makes it easier for Grad-CAM to localize. In contrast, COCO dataset contains images that describe sophisticated scenes, which can be challenging for Grad-CAM to identify all relevant regions. We would argue that the performance gain for the attention alignment model will increase by replacing Grad-CAM with better visualization algorithms. Also visual concepts from captions were generated using a small manually built word to category mapping. It's possible that some visual words haven't been identified and didn't contribution to the attention penalty. 

In the future work, we will investigate alternative optimization methods, which can update gradient-based attention maps and train LSTM in an alternating way. Instead of relying on the manually built word to categories mapping, we will explore different methods to generate it automatically, such as measuring the distance between word and categories in the word embedding space. Moreover, in this work we only utilized the bottom-up and top-down attention module, and we are going to apply and evaluate other commonly used attention mechanisms.